\documentclass[11pt,a4paper]{article}
\usepackage[hyperref]{acl2019}
\usepackage{times}
\usepackage{latexsym}
\usepackage{url}
\usepackage{helvet}
\usepackage{courier}
\usepackage{graphicx}
\usepackage{amsmath}
\usepackage{latexsym}
\usepackage{amssymb}
\usepackage{multirow}
\usepackage{amssymb}
\usepackage{amsmath}
\usepackage{setspace}
\usepackage{booktabs}
\usepackage{caption}
\usepackage{subcaption}
\usepackage{float}

\aclfinalcopy 

\title{Alleviate Exposure Bias in Sequence Prediction \\ with Recurrent Neural Networks}

\author{
  Liping Yuan, Jiangtao Feng, Xiaoqing Zheng, Xuanjing Huang \\
  Fudan University \\
  {\tt \{lpyuan19, fengjt16, zhengxq, xjhuang\}@fudan.edu.cn}
  }

\date{}

\begin{document}
\maketitle
\begin{abstract}
A popular strategy to train recurrent neural networks (RNNs), known as ``teacher forcing'' takes the ground truth as input at each time step and makes the later predictions partly conditioned on those inputs. 
Such training strategy impairs their ability to learn rich distributions over entire sequences because the chosen inputs hinders the gradients back-propagating to all previous states in an end-to-end manner.
We propose a fully  differentiable training algorithm for RNNs to better capture long-term dependencies by recovering the probability of the whole sequence. 
The key idea is that at each time step, the network takes as input a ``bundle'' of similar words predicted at the previous step instead of a single ground truth. 
The representations of these similar words forms a convex hull, which can be taken as a kind of regularization to the input.
Smoothing the inputs by this way makes the whole process trainable and differentiable.
This design makes it possible for the model to explore more feasible combinations (possibly unseen sequences), and can be interpreted as a computationally efficient approximation to the beam search. 
Experiments on multiple sequence generation tasks yield performance improvements, especially in sequence-level metrics, such as BLUE or ROUGE-2.

\end{abstract}

\section{Introduction}

Recurrent neural networks (RNNs) and their variants like long short-term memory (LSTM) have been empirically proven to be quite successful in structured prediction applications such as machine translation \cite{nips-sutskever:14}, Chatbot \cite{arxiv-vinyals:15}, parsing \cite{emnlp-ballesteros:16},  summarization \cite{emnlp-rash:15} and image caption generation \cite{cvpr-vinyals:15} due to their capability to bridge arbitrary time lags. 
By far the most popular strategy to train RNNs is via the maximum likelihood principle, which consists of maximizing the probability of the next word in the sequence given the current (recurrent) state and the previous ground truth word (also known as \emph{teacher forcing}). 
At inference time, truth previous words are unknown, and then are replaced by words predicted by the model itself.

\begin{figure}[t]
 \small
 \centering
 \includegraphics[width=7.5cm]{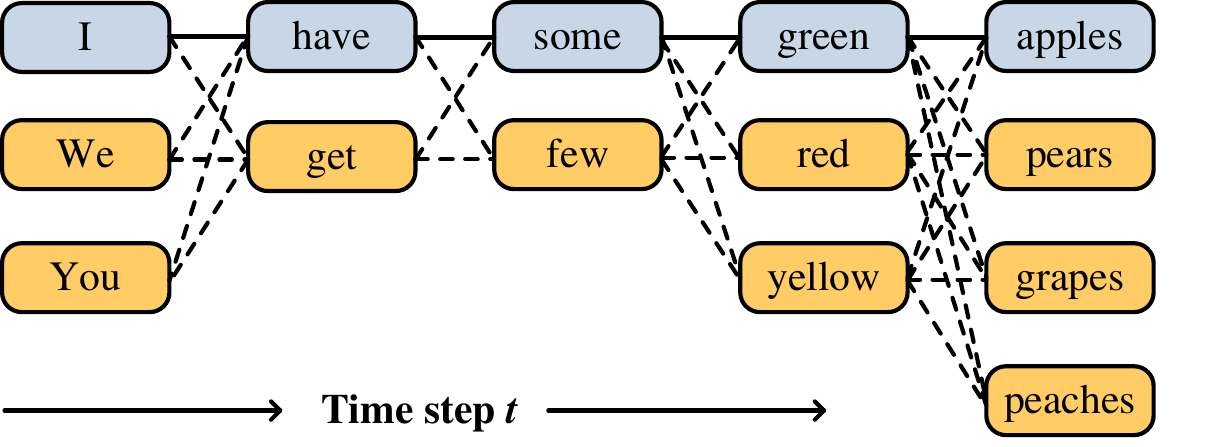}
 \caption{An example training sentence. For each time step, there are a single ground truth word (highlighted by grey color) and multiple semantically and syntactically similar words (highlighted by orange color). In our approach, the weighted embedding of those words at the time step $t$ is fed into the network at the next time step. Such design makes it possible for the model to explore more feasible combinations (e.g. ``I get few red graphs'' or ``We have some yellow peaches''), and can be considered as an approximation to the beam search in a less computationally demanding way.}
 \label{fig:sequence}
\end{figure}

The models trained by such teacher forcing strategy suffer from at least three drawbacks. 
First, the discrepancy between training and inference, called \emph{exposure bias} \cite{iclr-ranzato:16}, can yield errors because the model is only exposed to the distribution of training data, instead of its own predictions at inference. 
As a result the errors can accumulate quickly along the sequence to be generated. 
Second, the training loss for RNNs is usually defined at the word-level using maximum likelihood estimation (MLE), while their performances are typically evaluated using discrete and non-differentiable sequence-level metrics, such BLUE \cite{acl-papineni:02} or ROUGE \cite{acl-lin:04}. 
We call it \emph{evaluation bias}. 
Third, the whole training process is not \emph{fully differentiable} because the chosen input at each time step hinders the gradients back-propagating to all previous states in an end-to-end manner. 
Although Goodfellow et al \shortcite{book-goodfellow:16} pointed out that the underlying graphical model of these RNNs is a complete graph, and no conditional independence assumption is made to simplify the sequence prediction, the individual output probabilities might be still conditioned on the last few predictions if we use the teacher forcing regimen \cite{iclr-leblond:18}. 
Note that the chain rule can not be applied any longer where an incorrect prediction is replaced with its ground truth, which prevents the model from capturing long-term dependencies and recovering the probability of the whole sequence.

Many approaches have been proposed to deal with the first two drawbacks of training RNNs by scheduled sampling \cite{bengio2015scheduled}, adversarial domain adaptation \cite{nips-goyal:16}, reinforcement learning \cite{iclr-ranzato:16,iclr-bahdanau:17}, or learning to search (L2S) \cite{emnlp-wiseman:16,iclr-leblond:18} while little attention has been given to tackling the last one. 
A fully differentiable solution is capable of back-propagating the gradients through the entire sequence, which alleviates the discrepancy between training and inference by feeding RNNs with the same form of inputs, and gently bridges the gap between the training loss defined at each word and the evaluation metrics derived from whole sequence. 
Thanks to the chain rule for differentiation, the training loss of differentiable solution is indeed a sequence level loss that involves the joint probability of words in a sequence. 

We propose a fully differentiable training algorithm for RNNs to addresses the issues discussed above. 
The key idea is that at the time step $t+1$, the network takes as input a ``bundle'' of predicted words at the previous time step $t$ instead of a single ground truth. 
Intuitively, the input words at each time step need to be similar in terms of semantics and syntax, and they can be replaced each other without harming the structure of sentences (see Figure \ref{fig:sequence}).
The mixture of the similar words can be represented by a convex hull formed by their representations, which could be viewed as regularization to the input of recurrent neural networks.
This design makes it possible for the model to explore more feasible combinations (possibly unseen sequences), and can be interpreted as a computationally efficient approximation to the beam search.

We also want the number of input words can vary for different time steps, unlike the end-to-end algorithm proposed by Ranzato et al \shortcite{iclr-ranzato:16}, in which they propagated as input the predicted top-$k$ words, and the value of $k$ is fixed in advance.
Although different numbers of $k$ can be tested to determine the optimal value, the number of proper words that can be used heavily depends on contexts in which they occur. 
For a given $k$, we could introduce unnecessary noise if more words are taken as input, whereas we may prevent the model from exploring possible sequences if less words are involved. 
In our architecture, an attention mechanism \cite{nips-Vaswani:17} is applied to select candidate words, and their weighted embedding are fed into the network according to their attention scores. 
Smoothing the input by this way makes the whole process trainable and differentiable by using standard back-propagation. 


\section{Related Work}

Unlike Conditional Random Fields \cite{icml-Lafferty:01} and other models that assume independence between outputs at different time steps, RNNs and their vriants are capable of representing the probability distribution of sequences in the relatively general form. However, the most popular strategy for training RNNs, known as ``teacher forcing'' takes the ground truth as input at each time step, and makes the later predictions partly conditioned on those inputs being fed back into the model. Such training strategy impairs their ability to learn rich distributions over entire sequences. Although some training methods, such as scheduled sampling \cite{bengio2015scheduled} and ``professor forcing'' \cite{nips-goyal:16} are proposed to encourage the states of recurrent networks to be the same as possible when the training and inference over multiple time steps, the neural networks using greedy predictions (top-$1$) by the arguments of the maxima (abbreviated argmax) is not fully differentiable. Discrete categorical variables are involved in those greedy prediction, and become the obstacle to permit efficient computation of parameter gradients.

Perhaps the most natural and na\"{i}ve approach towards the differentiable training is that at time step $t + 1$ we take as input all the words whose contributions are weighted by their scores instead of the ground truth word, and the scores are the predicted distribution over words from the previous time step $t$. However, the output distribution at each time step is normally not sparse enough, and thus the input may blur by a large number of words semantically far from the ground truth. A simple remedy would be the end-to-end algorithm proposed by Ranzato et al \shortcite{iclr-ranzato:16}, in which they propagated as input the top $k$ words predicted at the previous time step. The $k$ largest scoring words are weighted by their re-normalized scores (summing to one). Although different numbers of $k$ can be tested to choose the optimal value, the value of $k$ is fixed at each time step. It would better to learn to determine a particular $k$ for each time step, depending on a specific context. For a fixed $k$, we would introduce unnecessary noise if too many words are involved, whereas we may prevent the model from exploring possible sequences if too few words are chosen.

Jang et al \shortcite{jang2017categorical} proposed Gumbel-Softmax that can make the output distributions over words more sparse by approximating discrete one-hot categorical distributions with their continuous analogues. Such approximate sampling mechanism for categorical variables can be integrated into RNNs, and trained using standard back-propagation by replacing categorical samples with Gumbel-Softmax ones. Theoretically, as the temperature $\tau$ of Gumbel-Softmax distribution approaches $0$, samples from this distribution will become one-hot. In practice, they begin with a high temperature and anneal to a small but non-zero one. The temperature, however, can not be too close to $0$ (usually $\tau \ge 0.5$) because the variance of the gradients will be so large, which may result in the gradient explosion.

Ideas coming from learning to search \cite{emnlp-wiseman:16,iclr-leblond:18} or reinforcement learning, such as the REINFORCE \cite{iclr-ranzato:16} and ACTOR-CRITIC \cite{iclr-bahdanau:17}, have been used to derive training losses that are more closely related to the sequence-level evaluation metrics that we actually want to optimize. Those approaches side step the issues associated with discrete nature of the optimization by not requiring losses to be differentiable. While those approaches appear to be well suited to tackle the training problem of RNNs, they suffer from a very large action space which makes them difficult to learn or search efficiently, and thus slow to train the models, especially for natural language processing tasks normally with a large vocabulary size. We reduce the effect of search error by pursuing few next word candidates at each time step in a pseudo-parallel way. We note that those ideas and ours are complementary to each other and can be combined to improve performance further, although the extent of the improvements will be task dependent.

\section{Methodology}

\begin{figure}[t]
    \centering
    \includegraphics[width=7.5cm]{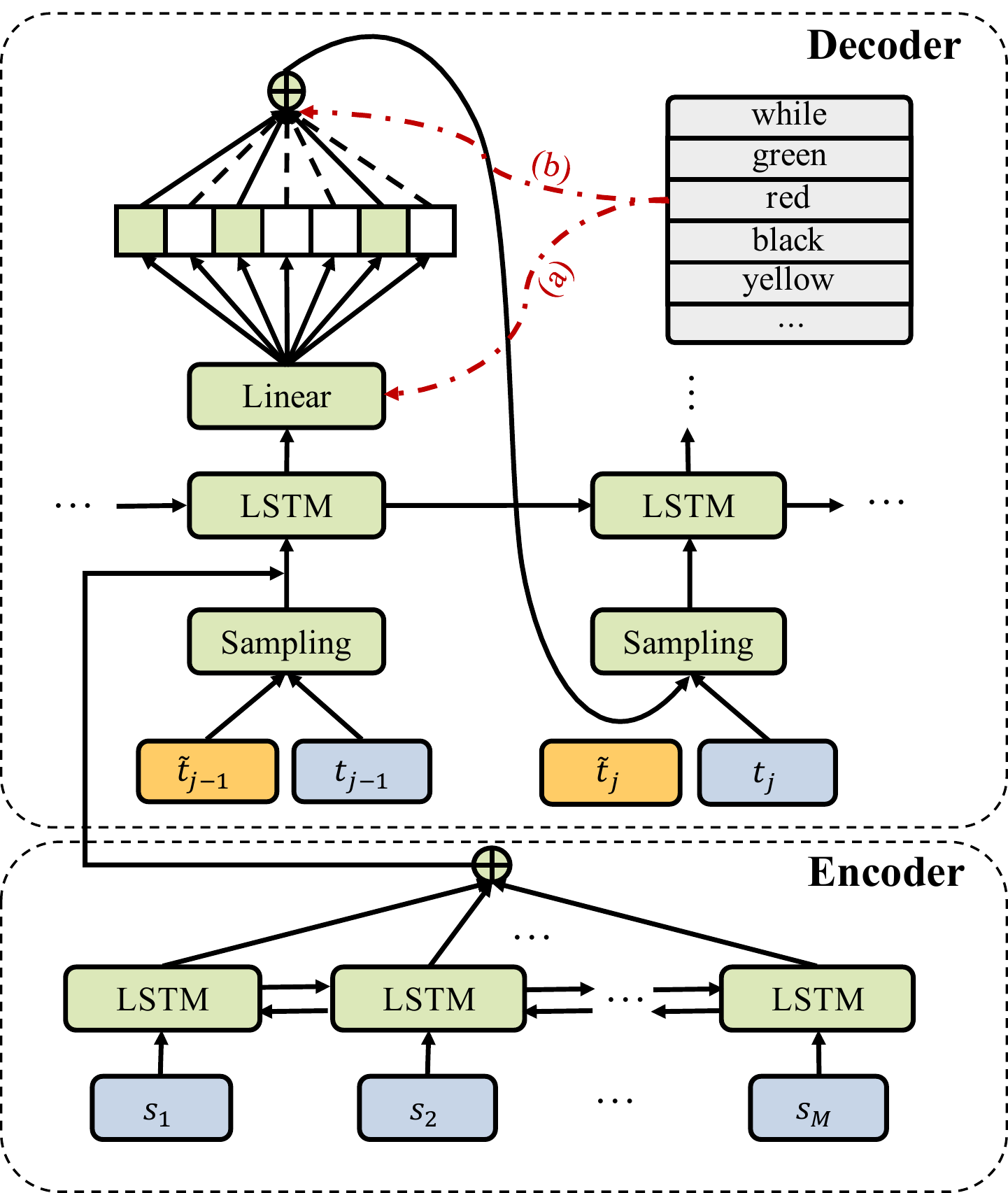}
    \caption{\small{M-WEAM architecture. It is based on a variational encoder-decoder with attention mechanism. (a) performs an attention operation over all the word embeddings to produce their probabilities (reflecting how likely they are chosen); (b) retrieves the words according to their probabilities (masked to make it sparse). By applying a margin relaxed method, we rule out the words having the lower probabilities than a threshold, and the remaining values are normalized to estimate the distribution of input words at the next time step.}}
    \label{fig:reg-model}
\end{figure}

We propose a fully differentiable method to train a sequence model, which allows the gradients to back-propagate through the entire sequence. 
Following \citep{zhang2016generating,jang2017categorical,goyal2017differentiable}, the network takes as input the weighted sum of word embeddings at each time step, where the weights reflect how likely they are chosen in the previous step. 
The following discussion is based on a general architecture that can be viewed as a variational encoder-decoder with an attention mechanism \citep{bahdanau2014neural, kingma2014auto-encoding, bowman2016generating}. 
The teacher forcing method, denoted as VAE-RNNSearch, is implemented with the above architecture, while we extend this architecture with a lexicon-based memory (Figure \ref{fig:reg-model} shows the details, where the margin relaxed component should be ignored at present), called \textbf{W}ord \textbf{E}mbedding \textbf{A}s \textbf{M}emory (WEAM).

Given a source sentence $s_{1:M}$, $s_i \in \mathcal{V}^s$ where $\mathcal{V}^s$ is the vocabulary of source language, a multi-layer BiLSTM encoder $f^s(\cdot)$ is used to compute a high-level contextual word representation $e_i \in \mathbb{R}^d$ for each word $s_i$, where $d$ is the dimensionality of the embedded vector space,
\begin{equation} \small
    e_{1:M} = f^s(s_{1:M})
\end{equation}
which is also the entries to be attended by the decoder. 
The final state of BiLSTM is used to compute the mean and the log variance of the latent Gaussian distribution $\mathcal{N}(\mu, \sigma^2)$ in VAE with a linear layer. 

In the decoding process, a seed $c$ sampled for the latent Gaussian distribution is fed into a multi-layer LSTM decoder $f^t(\cdot)$ with multi-hop attention. 
The decoder aims at predicting the distributions $p_{1:N}$ of target words $t_{1:N}$ conditioned on the seed $c$ and the encoder output $e_{1:M}$,
\begin{equation} \small
\label{eq:decoder}
    p_{1:N}=f^t(c, e_{1:M})
\end{equation}
where each $p_j\in [0, 1]^{|\mathcal{V}^t|}$ is the probability distribution of the $j$-th target word, and $\mathcal{V}^t$ is the vocabulary of target language.
The words with the maximum probability are the predictions $\tilde{t}_{1:N}$.

Specifically, we explain the decoding process in a recurrent manner. 
Above all, the word embedding matrix $\mathcal{M} \in \mathbb{R}^{|\mathcal{V}^t| \times d}$ is considered as a memory.
The recurrent function $g(\cdot)$ is a multi-layer LSTM cell incorporated with multi-hop attention over encoder output $e_{1:M}$. 
At timestamp $j$, the recurrent function $g(\cdot)$ output a hidden representation $h_{j+1}$ by giving the previous states $h_{j}$ and the current word embedding $v_j$, and the hidden state is further used to computed the likelihood of the predictions
\begin{equation} \small
    h_{j+1} = g(h_j, v_j)
\end{equation}
\begin{equation} \small
    p_{j+1} = \text{softmax}(\mathcal{M}h_{j+1})    
\end{equation}
For the starting timestamp, $h_0$ is defined as the seed $c$, and $v_0$ is the word embedding of the end-of-sentence token. 
During the training process, $v_j$ used in VAE-RNNSearch is the ground truth word embedding $\mathcal{M}(t_j)$ and that in WEAM is approximated by attention over word embeddings. 
The attention score here could be obtained by reusing the predicted distribution of the next word, namely $p_j(\cdot)$. 
Thus the input word embedding $v_j$ is approximated by
\begin{equation} \small
\label{eq:embed_atten}
    v_j = p_j\mathcal{M} = \sum_{w\in\mathcal{V}^t}{p_j(w)\mathcal{M}(w)}
\end{equation}
From now on, we use ground truth words to denote words inputted into the network and target words to denote objective ones for the purpose of distinguishing their functions, although they refer to the same sequence.

Finally, after the entire sequence is generated, the objective function of WEAM is computed from the predicted probability $p_j(\cdot)$ and the target sequence $t_{1:N}$ as well as the latent Gaussian distribution, 
\begin{equation} \small
\begin{split}
\label{eq:loss}
    L(\theta) = & -\sum_{j=1..N}{\log p_j(t_j)} \\
                & + D_{KL}(\mathcal{N}(\mu, \sigma^2) || \mathcal{N}(0,1))
\end{split}
\end{equation}
In Equation \ref{eq:loss}, the first term is the cross-entropy between the predicted probability with the target one, and the second term is the KL divergence between the approximate latent Gaussian distribution and a prior normal distribution. 
Equipped with WEAM, the training process of text generation is fully differentiable.

At the inference, instead of using the attention-weighted embedding in Equation \ref{eq:embed_atten}, we found it better to feed an exact word embedding into the neural network at each timestamp. 
Although the word embedding memory regularizes $v_{j+1}$ in a convex hull of word embeddings comparing with $h_j$ (Figure \ref{fig:reg-weam}), there is still difference with the representations of genuine words in vocabulary, which is a finite set.
The bias could lead to minor mistakes, which accumulate as the prediction process goes on and is potentially harmful to future predictions.
Thus feeding an exact word embedding in inference would be helpful in generating a sequence by regularizing the input embedding and erasing the potential mistakes.

\subsection{Sparse Word Distributions}

As discussed above, feeding an exact word embedding is a helpful property in inference, which is equivalent to setting a sparse distribution $p_j(\cdot)$.
However, WEAM model suffers from the long tail effect cause by the non-sparse $p_j(\cdot)$.
Most words in vocabulary are predicted with a very small but non-negligible probability, leading to a noisy $v_j$.

A widely-used method to obtain a sparse distribution is rescaling the logits before presenting it to the final-layer classifier, equivalently
\begin{equation} \small
\label{eq:rescaled-logit}
    \tilde{p}_j(k) = p_j(k)^{1/\tau}
\end{equation}
where $\tau$ is a rescaling factor decaying from one to zero \citep{zhang2016generating, jang2017categorical}. 
Ideally, when $\tau$ converges to zeros, the distribution $\tilde{p}_j(\cdot)$ converges to a categorical one.
However, in practice, $\tau$ cannot be set to a small value, which could lead to gradient explosion. 
Thus an alternative strategy is setting a lowerbound for $\tau$, which is empirically $0.5$ in \citep{jang2017categorical}. 
This value of $\tau$ is far from the optimal number (i.e. $0$) in theory, and the sparse categorical distribution can be achieved in practice. 

\begin{figure}[t]
    \centering
    \begin{subfigure}{0.23\textwidth}
        \includegraphics[width=\textwidth]{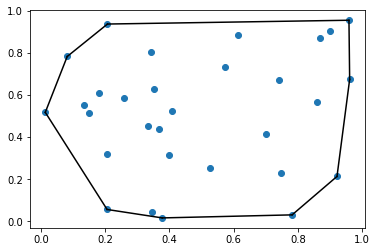}
        \subcaption{WEAM}
        \label{fig:reg-weam}
    \end{subfigure}
    \begin{subfigure}{0.23\textwidth}
        \includegraphics[width=\textwidth]{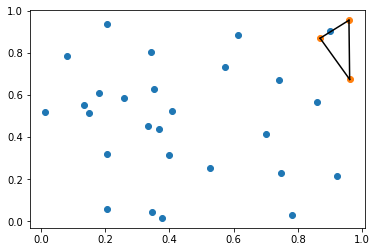}
        \subcaption{M-WEAM}
        \label{fig:reg-rweam}
    \end{subfigure}
    \caption{The difference between (a) WEAM and (b) M-WEAM in the way they form the subspace in a convex hull defined by all the word embeddings, which can be viewed as a kind of regularizations. Blue points represent all the words in a vocabulary, whereas yellow ones are the unmasked words in M-WEAM. The simplex is the convex hull defined by the unmasked words chosen by the results of  the lexicon-based attention.}
\label{fig:reg}
\end{figure}

We propose a margin relaxed method to make a sparse distribution through restricting $v_j$ in a relatively smaller subspace (see Figure \ref{fig:reg-rweam}). 
The architecture of M-WEAM model is shown in Figure \ref{fig:reg-model}. 
Given a distribution $p_j(\cdot)$, its sparse version $\tilde{p}_j(\cdot)$ is computed by ruling out the words with low probability,
\begin{equation} \small
    \tilde{p}_j(w) = \frac{p_j(w)mask_j(w)}{\sum_{l}{p_j(w')mask_j(w')}}
\end{equation}
\begin{equation} \small
    mask_j(w) = \begin{cases} 1 & p_j(w) > \eta_j \\
    0 & otherwise
    \end{cases}
\end{equation}
where $\eta_j$ is a threshold and could be determined by various ways.
Its motivation is to alleviate the long tail effect by masking the ``long tail'' entries with close-to-zero probability.
A possible choice is defined the threshold $\eta_j$ as 
\begin{equation} \small
\label{eq:margin}
    \eta_j=e^{-\epsilon}\max_{k}p_j(k)
\end{equation}
where $\epsilon$ is a margin in log space.
As $\epsilon$ converges to $0$, the $\tilde{p}_j(k)$ converges to a categorical distribution as well.
However, we do not intend to anneal $\epsilon$ to zero, aiming at keep a mixture meaning of multiple semantically and syntactically similar words in the training process.
In vector space, the mixture meaning is represented as the convex hull shown in Figure \ref{fig:reg-rweam}.
All the words forming this convex hull are reasonable predictions, and their representations are optimized in the training.
If $\epsilon$ become very small, there is only one word, namely one point, to be optimized, leading to slower convergence.
Besides, an input representation $v_j$ generated from a subspace would be more robust than that assigned to only one point, preventing the model from overfitting to a specific point.
The optimization over a subspace could be considered as a balance between that over a point and over the whole space in Figure \ref{fig:reg-weam}.

The re-estimated probability $\tilde{p}(\cdot)$ can also be used in objective function Equation \ref{eq:loss} following ``enough is enough'' principal.
Note that the training objective is discriminating a target word with the others in vocabulary by increasing the margin between them.
As training progresses, the margin becomes larger implying that the target word can be easily pointed out from the vocabulary. 
Keeping optimizing the margin between the target word and all the others would be useless, and easily leads to overfitting.
The ``enough is enough'' principal means that a word requires no further optimization when its likelihood is much lower than the target one. 
Thus the re-estimated probability $\tilde{p}(\cdot)$ is used in the loss function to prevent the words with low probability from over-optimization.

We denote the vanilla non-sparse version as WEAM, the one rescaling logits in Equation \ref{eq:rescaled-logit} as R-WEAM, and the proposed margin relaxed one as M-WEAM. Note that R-WEAM is slightly different from that in \citep{zhang2016generating, jang2017categorical} since word classifier and word embeddings are tied \cite{inan2017tying}. 

\subsection{Warm-up Strategy}

At the beginning of the training, the words generated by WEAM models are almost random. 
Conditioned on such randomly predicted words, the model fails to benefit from the future training.
In order to alleviate this problem, the models are warmed up using the teacher forcing strategy. We present two warming up techniques to train WEAM-family models.

\subsubsection{Scheduled Sampling}

Scheduled sampling is a widely-used way to warm up WEAM-family models in training \citep{bengio2015scheduled}.
Whether to use the embedding of a ground truth word or its approximation in Equation \ref{eq:embed_atten} at the next timestamp is randomly selected as:
\begin{equation} \small \label{eq:sample-i}
\begin{split}
    v_{j+1} & =\begin{cases}
        \mathcal{M}(w_{j+1}), & t=0 \\
        \sum_{w\in\mathcal{V}^t}{p_j(w)M(w)}, & t=1
    \end{cases} \\
    t & \sim \mathcal{B}(\text{steps}/\text{max\_steps})
\end{split}
\end{equation}
where $t$ is sampled from a Bernoulli distribution $\mathcal{B}(\cdot)$ whose expectation has a positive correlation with training progress.
From a training perspective, the probability of choosing the approximated word embedding increases linearly from zero to one.

\subsubsection{Threshold}

Masking words with low probability may cause a new problem that a target word may be masked and lose its opportunity for optimization.
Once a target word is ruled out of training, it may be ruled out forever since its low probability, leading to bad performances and slow convergence.
An alternative way to solve this problem is computing the threshold with the probability of the target word, and we determines the threshold in a probable way as:
\begin{equation} \small \label{eq:gold-margin}
\begin{split}
    \eta_j & =\begin{cases}
    e^{-\epsilon} p_j(w_{j+1}), & r=0 \\
    e^{-\epsilon} \max_{k}e^{-\epsilon}p_j(k), & r=1
    \end{cases} \\
    r & \sim \mathcal{B}(\max(\text{steps}/\text{max\_steps}, \xi))
\end{split}
\end{equation}
where $r$ is similar to $t$ with lowerbound $\xi$.
Equation \ref{eq:gold-margin} shows that the threshold is determined by the target word at the beginning of training whereas it is determined by the predictions when the algorithm almost converges.
Adopting a lowerbound $\xi$ also leads to an attractive property that a prediction and its re sptteiaevcrget word are replaceable to each other, and thus the model is expected to capturing the phenomenon of synonyms.

\section{Experiments}

We conducted three sets of experiments to demonstrate the effectiveness of M-WEAM on various tasks including machine translation, text summarization and dialogue generation. 

We employed almost the same setting to all the tasks. 
A wrapped recurrent block was used for both the encoder and decoder, defined as a sequence of LSTM, dropout, residual connection and layer normalization. 
Uni-directional LSTM was used in encoder while the bi-directional LSTM was used in decoder.
The multi-hop attention component was also wrapped in a similar way.
We set the dimensionality of vector space to $256$, layers of recurrent block to $3$, dropout rate to $0.3$, lowerbound of margin $\epsilon$ to $1$, and $\xi$ to $0.5$.
The models were optimized by Adam \cite{kingma2014adam} with $\beta_1=0.9, \beta_2=0.999 $ and $\text{eps}=10^{-8}$. 
Learning rate was scheduled following \cite{feng2018neural}, and its maximum was set to $0.001$.
Label smoothing was also used with $0.1$ smoothing rate. 
Separated vocabularies were constructed for source and target domains on translation and dialogue tasks whereas a shared one was used for summarization task.
All the models were implemented with PyTorch and trained on $1$ NVIDIA Titan Xp GPU.


\subsection{Machine Translation}

\begin{table}[b]
    \centering
    \begin{tabular}{l|c}
        \toprule
        Model & BLEU \\
        \midrule
        MIXER & $20.73$ \\
        BSO & $23.83$ \\
        $\alpha$-soft annealing & $20.60$ \\
        Actor-Critic & $27.49$ \\
        SEARNN & $28.20$ \\
        NPMT & $28.57$ \\
        \midrule
        VAE-RNNSearch & $28.97$ \\
        WEAM & $27.84$ \\
        R-WEAM & $28.13$\\
        \midrule
        M-WEAM & \textbf{29.17} \\
        \bottomrule
    \end{tabular}
    \caption{Results on IWSLT14 German-English.}
    \label{tab:de-en}
\end{table}

Machine translation task was experimented on two datasets, IWSLT14 German to English (De-En) \citep{cettolo2014report} and IWSLT15 English to Vietnamese (En-Vi) \citep{cettolo2015iwslt}. 
The dataset was preprocessed basically following \cite{huang2017towards, feng2018neural}. 
For IWSLT14 De-En, the dataset roughly contains 153K training sentences, 7K development sentences and 7K testing sentences. 
Words whose frequency of occurrence is lower than 3 were replaced by an ``UNK'' token.
For IWSLT15 En-Vi, the dataset contains 133K translation pairs in training set, 1,553 in validation set (TED tst2012) and 1,268 in test set (TED tst2013).
Similar to De-En, words with frequency of occurrence lower than 5 were replaced by the ``UNK'' token.

\begin{table}[t]
    \centering
    \begin{tabular}{l|c}
         \toprule
         Model & BLEU \\
         \midrule
         Hard Monotonic & $23.00$ \\
         NPMT & $26.91$ \\
         \midrule
         VAE-RNNSearch & $28.66$ \\
         WEAM & $27.14$ \\
         R-WEAM & $26.79$ \\
         \midrule
         M-WEAM & \textbf{28.70} \\
         \bottomrule
    \end{tabular}
    \caption{Results on IWSLT15 English-Vietnamese.}
    \label{tab:en-vi}
\end{table}

On IWSLT14 De-En, we compared M-WEAM against VAE-RNNSearch \citep{bowman2016generating}, WEAM, R-WEAM \citep{zhang2016generating}, MIXER \citep{iclr-ranzato:16}, BSO \citep{emnlp-wiseman:16}, Actor-Critic \citep{iclr-bahdanau:17}, SEARNN \citep{iclr-leblond:18}, $\alpha$-soft annealing \cite{goyal2017differentiable}, and NPMT \citep{huang2017towards}.
On IWSLT15 En-Vi, we compared M-WEAM against VAE-RNNSearch \citep{bowman2016generating}, WEAM, R-WEAM \citep{zhang2016generating}, Hard Monotonic \citep{raffel2017online} and NPMT \citep{huang2017towards}  .


Table \ref{tab:de-en} and Table \ref{tab:en-vi} display the results on the German-English and English-Vietnamese translation task respectively. 
It is can be seen that M-WEAM achieves decent results on the both tasks.
Comparing with differentiable models, M-WEAM seizes a notable gain of over $1$ BLEU point, outperforming WEAM and R-WEAM on both tasks. 
M-WEAM also achieves comparable results with its teacher forcing version.
Besides, M-WEAM beats previous SOTA RNN-based models with a significant margin.

Inspired by \cite{keskar2016large}, where they explored the neighborhood of the minima to measure the model's sensitivity, we conducted a similar experiment on IWSLT14 De-En to show the robustness of VAE-RNNSearch and WEAM family models. 
We tested on the models by injecting noise into input embedding. 
The noise was sampled from an unbiased Gaussian distribution with the standard deviation increasing from 0 to 1. 
As the noise increases, models with sharper local minima should suffer more from performance degradation.  

\begin{figure}[t]
    \centering
    \includegraphics[width=7.5cm]{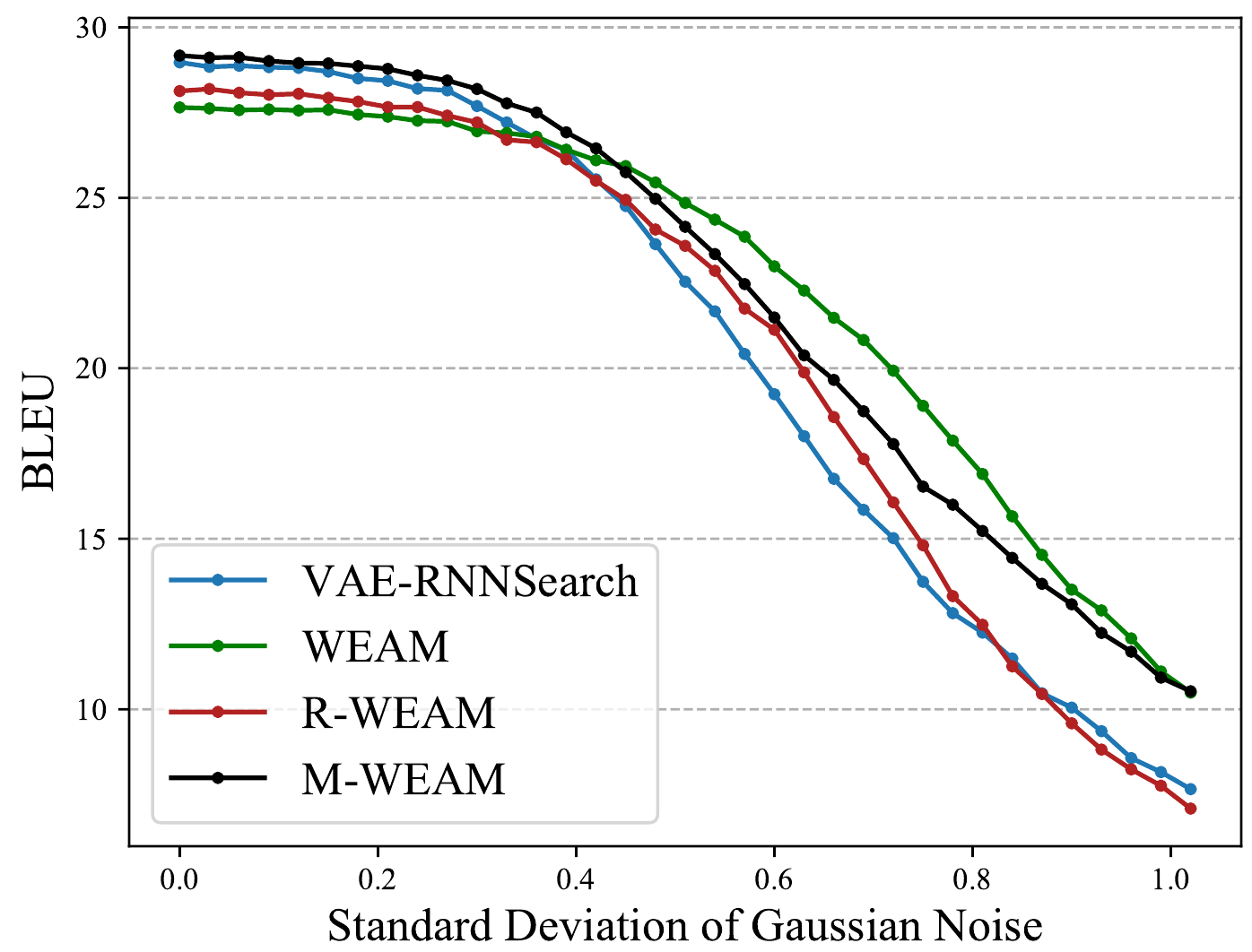}
    \caption{Sensitivity test with different Gaussian noise levels. The $x$-axis is the standard deviation of the Gaussian noise. 
    }
    \label{fig:sensitivity}
\end{figure}

Figure \ref{fig:sensitivity} shows how the performance of the compared models changes against noise. 
M-WEAM model surpasses the teacher forcing model by nearly 3 BLEU points when the standard deviation reaches 1. 
It thus can be inferred that M-WEAM converges to a flatter minima than VAE-RNNSearch does.
Besides, among all models, WEAM shows strongest anti-noise ability because it has been trained with massive noise due to the long tail effect. 
In general, M-WEAM achieves a balance between performance and robustness.

\subsection{Abstractive Summarization}

We evaluated the models on Gigaword benchmark, and applied the same data preprocessing as \citep{rush2015neural, chopra2016abstractive}.
The results are reported with F1 ROUGE scores, including ROUGE-1, ROUGE-2 and ROUGE-L.

The results of ABS \citep{rush2015neural}, ABS+ \citep{rush2015neural}, Feats \citep{nallapati2016abstractive}, RAS-LSTM \citep{chopra2016abstractive} and RAS-Elman \citep{chopra2016abstractive} are extracted from the numbers reported by their authors. We implemented the VAE-RNNSearch \citep{bowman2016generating} as well as WEAM, and report their results using our implementations. The results of ABS and ABS+ with the greedy search are unavailable, and thus we report those with the beam search instead. We are aware that SEASS \citep{zhou2017selective} and DRGD \citep{li2017deep} are two recent studies in which the higher ROUGE metrics were reported on Gigaword dataset. 
Their results are not listed because those models are specifically tailored for the abstractive summarization task. SEASS is equipped with a selective network for salient language units, and DRGD uses a recurrent latent variable to capture the summary structure, while we focus on the general framework and training algorithm for the sequence-to-sequence tasks. 


\begin{table}[t]
    \centering
    \begin{tabular}{l|c|c|c}
        \toprule
        Model & R-1 & R-2 & R-L \\
        \midrule
        ABS$^*$ & $29.55$ & $11.32$ & $26.42$ \\
        ABS+$^*$ & $29.78$ & $11.89$ & $26.97$ \\
        Feats & $32.67$ & $15.59$ & $30.64$ \\
        RAS-LSTM & $31.71$ & $13.63$ & $29.31$ \\
        RAS-Elman & $33.10$ & $14.45$ & $30.25$ \\
        \midrule
        VAE-RNNSearch & $\textbf{33.99}$ & $15.72$ & $\textbf{31.67}$ \\
        WEAM & $33.09$ & $15.05$ & $30.79$ \\
        R-WEAM & $32.35$ & $14.52$ & $30.10$ \\
        \midrule
        M-WEAM & $33.87$ & $\textbf{15.78}$ & $31.52$ \\
        \bottomrule
    \end{tabular}
    \caption{Results on Gigaword dataset. The models indicated with $^*$ use beam search to generate summaries whereas the others use the greedy search instead.}
    \label{tab:gigaword}
\end{table}

Table \ref{tab:gigaword} shows the results on Gigaword Test Set. 
Among the fully differentiable WEAM family models, M-WEAM achieves the best performance, and outperform the WEAM and R-WEAM with at least $0.7$ improvement on the all ROUGE metrics.
By comparing with VAE-RNNSearch, M-WEAM achieves comparable results on all the ROUGE metrics. 
It is worth noting that although M-WEAM performs slightly worse than VAE-RNNSearch on ROUGE-1, but it does better on ROUGE-2, which shows that M-WEAM is able to recover more longer patterns, thanks to its fully differentiable property.

\subsection{Dialogue}

Finally, we evaluated the models on the task of single-turn dialogue, and used a subset of STC-weibo corpus \cite{shang2015neural} as a dataset, where weibo is a popular Twitter-like microblogging service in China. 
The complete corpus consists of roughly $4.4$M conversational post-response pairs.
We removed the sentences containing more than $10$ words and created a dataset with $907,809$ post-response pairs for fast training and testing.
The dataset was randomly split into training, validation, and testing sets, which contains $899$K, $10$K and $10$K pairs respectively. 
The vocabulary was formed by the most frequent $40$K words in training data.

BLEU score is a widely used metric in dialogue evaluation to measure the response relevance, whereas distinct-1/2 is used to measure the response diversity. 
As is shown in Table \ref{tab:stc}, M-WEAM model achieves the highest performance in terms of BLEU, but performs worse than other models with distinct metrics. 
Unlike machine translation, a response can be taken as an acceptable answer to many posts (or questions) and the dataset contains a number of one-to-many cases. In order to recover the probability of entire sequences, the fully differentiable model with improved training efficiency tends to generate frequently-used sequences and $n$-grams, which hurts the diversity of the generated responses. We leave this issue for future research. 
The high relevance but low diversity of responses generated by M-WEAM show that the proposed M-WEAM model is more desirable for the tasks requiring to generate exact sequences.

\begin{table}[t]  
    \small
    \centering
    \begin{tabular}{l|c|c|c}
        \toprule
        Model & BLEU & distinct-1 & distinct-2 \\
        \midrule
        VAE-RNNSearch & $5.95$ & $2.39$ & $19.01$ \\
        WEAM & $5.67$ & \textbf{2.50} & $19.52$ \\
        R-WEAM & $5.73$ & $2.49$ & \textbf{19.65} \\
        \midrule
        M-WEAM & \textbf{6.11} & $2.39$ & $16.31$ \\
        \bottomrule
    \end{tabular}
    \normalsize
    \caption{Automatic evaluation results on STC-weibo dataset. The best results are highlighted in bold font.}
    \label{tab:stc}
\end{table}

\section{Conclusion}

We proposed a fully differentiable training algorithm for RNNs to alleviate the discrepancy between training and inference (exposure bias), and to bridge the gap between the training loss defined at each word and the evaluation metrics derived from whole sequence (evaluation bias). 
The fully differentiable property is achieved by feeding the network at each step with a ``bundle'' of the words carrying similar meaning instead of a single ground truth. 
In our solution, the network is allowed to take different number of words as input at each time step, depending on the context. 
We may have more candidate words at some positions, while less at others when trying to generate a sentence. 
Experiments on machine translation, abstractive summarization, and open-domain dialogue response generation tasks showed that the proposed architecture and training algorithm achieved the best or comparable performance, especially in BLUE or ROUGE-2 metrics defined on sequence level, reflecting the enhanced ability in capturing long-term dependencies and recovering the probability of the whole sequence. 
The trained networks is also empirically proven to be more robust to noises, and perform constantly better other competitors with different noise levels.


\bibliography{ACL2019}

\begin{thebibliography}{36}
\expandafter\ifx\csname natexlab\endcsname\relax\def\natexlab#1{#1}\fi

\bibitem[{Alexander M.~Rush(2015)}]{emnlp-rash:15}
Jason~Weston Alexander M.~Rush, Sumit~Chopra. 2015.
\newblock A neural attention model for sentence summarization.
\newblock In \emph{Proceedings of the Conference on Empirical Methods in
  Natural Language Processing}.

\bibitem[{Bahdanau et~al.(2017)Bahdanau, Brakel, Xu, Goyal, Lowe, Pineau,
  Courville, and Bengio}]{iclr-bahdanau:17}
Dzmitry Bahdanau, Philemon Brakel, Kelvin Xu, Anirudh Goyal, Ryan Lowe, Joelle
  Pineau, Aaron~C. Courville, and Yoshua Bengio. 2017.
\newblock \href {https://openreview.net/forum?id=SJDaqqveg} {An actor-critic
  algorithm for sequence prediction}.
\newblock In \emph{5th International Conference on Learning Representations,
  {ICLR} 2017, Toulon, France, April 24-26, 2017, Conference Track
  Proceedings}. OpenReview.net.

\bibitem[{Bahdanau et~al.(2014)Bahdanau, Cho, and Bengio}]{bahdanau2014neural}
Dzmitry Bahdanau, Kyunghyun Cho, and Yoshua Bengio. 2014.
\newblock Neural machine translation by jointly learning to align and
  translate.
\newblock \emph{arXiv preprint arXiv:1409.0473}.

\bibitem[{Ballesteros et~al.(2016)Ballesteros, Goldberg, Dyer, and
  Smith}]{emnlp-ballesteros:16}
Miguel Ballesteros, Yoav Goldberg, Chris Dyer, and Noah~A Smith. 2016.
\newblock Training with exploration improves a greedy stack-\uppercase{LSTM}
  parser.
\newblock In \emph{Proceedings of the Conference on Empirical Methods in
  Natural Language Processing}.

\bibitem[{Bengio et~al.(2015)Bengio, Vinyals, Jaitly, and
  Shazeer}]{bengio2015scheduled}
Samy Bengio, Oriol Vinyals, Navdeep Jaitly, and Noam Shazeer. 2015.
\newblock \href
  {https://proceedings.neurips.cc/paper/2015/hash/e995f98d56967d946471af29d7bf99f1-Abstract.html}
  {Scheduled sampling for sequence prediction with recurrent neural networks}.
\newblock In \emph{Advances in Neural Information Processing Systems 28: Annual
  Conference on Neural Information Processing Systems 2015, December 7-12,
  2015, Montreal, Quebec, Canada}, pages 1171--1179.

\bibitem[{Bowman et~al.(2016)Bowman, Vilnis, Vinyals, Dai, Jozefowicz, and
  Bengio}]{bowman2016generating}
Samuel~R Bowman, Luke Vilnis, Oriol Vinyals, Andrew~M Dai, Rafal Jozefowicz,
  and Samy Bengio. 2016.
\newblock Generating sentences from a continuous space.
\newblock In \emph{Proceedings of the Conference on Computational Natural
  Language Learning}, pages 10--21.

\bibitem[{Cettolo et~al.(2015)Cettolo, Niehues, St{\"u}ker, Bentivogli,
  Cattoni, and Federico}]{cettolo2015iwslt}
Mauro Cettolo, Jan Niehues, Sebastian St{\"u}ker, Luisa Bentivogli, Roldano
  Cattoni, and Marcello Federico. 2015.
\newblock The iwslt 2015 evaluation campaign.
\newblock In \emph{IWSLT 2015, International Workshop on Spoken Language
  Translation}.

\bibitem[{Cettolo et~al.(2014)Cettolo, Niehues, St{\"u}ker, Bentivogli, and
  Federico}]{cettolo2014report}
Mauro Cettolo, Jan Niehues, Sebastian St{\"u}ker, Luisa Bentivogli, and
  Marcello Federico. 2014.
\newblock Report on the 11th iwslt evaluation campaign, iwslt 2014.
\newblock In \emph{Proceedings of the International Workshop on Spoken Language
  Translation, Hanoi, Vietnam}.

\bibitem[{Chopra et~al.(2016)Chopra, Auli, and Rush}]{chopra2016abstractive}
Sumit Chopra, Michael Auli, and Alexander~M Rush. 2016.
\newblock Abstractive sentence summarization with attentive recurrent neural
  networks.
\newblock In \emph{Proceedings of the 2016 Conference of the North American
  Chapter of the Association for Computational Linguistics: Human Language
  Technologies}, pages 93--98.

\bibitem[{Feng et~al.(2018)Feng, Kong, Huang, Wang, Huang, Mao, Qiao, and
  Zhou}]{feng2018neural}
Jiangtao Feng, Lingpeng Kong, Po{-}Sen Huang, Chong Wang, Da~Huang, Jiayuan
  Mao, Kan Qiao, and Dengyong Zhou. 2018.
\newblock Neural phrase-to-phrase machine translation.
\newblock \emph{arXiv preprint arXiv:1811.02172}.

\bibitem[{Goyal et~al.(2016)Goyal, Lamb, Zhang, Zhang, Courville, and
  Bengio}]{nips-goyal:16}
Anirudh Goyal, Alex Lamb, Ying Zhang, Saizheng Zhang, Aaron~C. Courville, and
  Yoshua Bengio. 2016.
\newblock \href
  {https://proceedings.neurips.cc/paper/2016/hash/16026d60ff9b54410b3435b403afd226-Abstract.html}
  {Professor forcing: {A} new algorithm for training recurrent networks}.
\newblock In \emph{Advances in Neural Information Processing Systems 29: Annual
  Conference on Neural Information Processing Systems 2016, December 5-10,
  2016, Barcelona, Spain}, pages 4601--4609.

\bibitem[{Goyal et~al.(2017)Goyal, Dyer, and
  Berg-Kirkpatrick}]{goyal2017differentiable}
Kartik Goyal, Chris Dyer, and Taylor Berg-Kirkpatrick. 2017.
\newblock \href {https://doi.org/10.18653/v1/P17-2058} {Differentiable
  scheduled sampling for credit assignment}.
\newblock In \emph{Proceedings of the 55th Annual Meeting of the Association
  for Computational Linguistics (Volume 2: Short Papers)}, pages 366--371,
  Vancouver, Canada. Association for Computational Linguistics.

\bibitem[{Huang et~al.(2018)Huang, Wang, Huang, Zhou, and
  Deng}]{huang2017towards}
Po{-}Sen Huang, Chong Wang, Sitao Huang, Dengyong Zhou, and Li~Deng. 2018.
\newblock \href {https://openreview.net/forum?id=HktJec1RZ} {Towards neural
  phrase-based machine translation}.
\newblock In \emph{6th International Conference on Learning Representations,
  {ICLR} 2018, Vancouver, BC, Canada, April 30 - May 3, 2018, Conference Track
  Proceedings}. OpenReview.net.

\bibitem[{Inan et~al.(2017)Inan, Khosravi, and Socher}]{inan2017tying}
Hakan Inan, Khashayar Khosravi, and Richard Socher. 2017.
\newblock Tying word vectors and word classifiers: A loss framework for
  language modeling.
\newblock \emph{international conference on learning representations}.

\bibitem[{Jang et~al.(2017)Jang, Gu, and Poole}]{jang2017categorical}
Eric Jang, Shixiang Gu, and Ben Poole. 2017.
\newblock \href {https://openreview.net/forum?id=rkE3y85ee} {Categorical
  reparameterization with gumbel-softmax}.
\newblock In \emph{5th International Conference on Learning Representations,
  {ICLR} 2017, Toulon, France, April 24-26, 2017, Conference Track
  Proceedings}. OpenReview.net.

\bibitem[{Keskar et~al.(2017)Keskar, Mudigere, Nocedal, Smelyanskiy, and
  Tang}]{keskar2016large}
Nitish~Shirish Keskar, Dheevatsa Mudigere, Jorge Nocedal, Mikhail Smelyanskiy,
  and Ping Tak~Peter Tang. 2017.
\newblock \href {https://openreview.net/forum?id=H1oyRlYgg} {On large-batch
  training for deep learning: Generalization gap and sharp minima}.
\newblock In \emph{5th International Conference on Learning Representations,
  {ICLR} 2017, Toulon, France, April 24-26, 2017, Conference Track
  Proceedings}. OpenReview.net.

\bibitem[{Kingma and Ba(2015)}]{kingma2014adam}
Diederik~P. Kingma and Jimmy Ba. 2015.
\newblock \href {http://arxiv.org/abs/1412.6980} {Adam: {A} method for
  stochastic optimization}.
\newblock In \emph{3rd International Conference on Learning Representations,
  {ICLR} 2015, San Diego, CA, USA, May 7-9, 2015, Conference Track
  Proceedings}.

\bibitem[{Kingma and Welling(2014)}]{kingma2014auto-encoding}
Diederik~P. Kingma and Max Welling. 2014.
\newblock \href {http://arxiv.org/abs/1312.6114} {Auto-encoding variational
  bayes}.
\newblock In \emph{2nd International Conference on Learning Representations,
  {ICLR} 2014, Banff, AB, Canada, April 14-16, 2014, Conference Track
  Proceedings}.

\bibitem[{Lafferty et~al.(2001)Lafferty, McCallum, and
  Pereira}]{icml-Lafferty:01}
John~D. Lafferty, Andrew McCallum, and Fernando C.~N. Pereira. 2001.
\newblock Conditional random fields: Probabilistic models for segmenting and
  labeling sequence data.
\newblock In \emph{Proceedings of the Eighteenth International Conference on
  Machine Learning {(ICML} 2001), Williams College, Williamstown, MA, USA, June
  28 - July 1, 2001}, pages 282--289. Morgan Kaufmann.

\bibitem[{Leblond et~al.(2018)Leblond, Alayrac, Osokin, and
  Lacoste-Julien}]{iclr-leblond:18}
R{\'e}mi Leblond, Jean-Baptiste Alayrac, Anton Osokin, and Simon
  Lacoste-Julien. 2018.
\newblock \uppercase{SEARNN}: Training \uppercase{RNN}s with global-local
  losses.
\newblock In \emph{Proceedings of the International Conference on Learning
  Representations}.

\bibitem[{Li et~al.(2017)Li, Lam, Bing, and Wang}]{li2017deep}
Piji Li, Wai Lam, Lidong Bing, and Zihao Wang. 2017.
\newblock Deep recurrent generative decoder for abstractive text summarization.
\newblock \emph{empirical methods in natural language processing}, pages
  2091--2100.

\bibitem[{Lin(2004)}]{acl-lin:04}
Chin-Yew Lin. 2004.
\newblock \uppercase{ROUGE}: A package for automatic evaluation of summaries.
\newblock In \emph{Proceedings of the Workshop on Text Summarization Branches
  Out, the Annual Meeting of the Association for Computational Linguistics}.

\bibitem[{Nallapati et~al.(2016)Nallapati, Zhou, Gulcehre, Xiang
  et~al.}]{nallapati2016abstractive}
Ramesh Nallapati, Bowen Zhou, Caglar Gulcehre, Bing Xiang, et~al. 2016.
\newblock Abstractive text summarization using sequence-to-sequence rnns and
  beyond.
\newblock In \emph{Proceedings of The 20th {SIGNLL} Conference on Computational
  Natural Language Learning}.

\bibitem[{Oriol~Vinyals(2015)}]{arxiv-vinyals:15}
Quoc V.~Le Oriol~Vinyals. 2015.
\newblock A neural conversational model.
\newblock \emph{arXiv:1506.05869}.

\bibitem[{Papineni et~al.(2002)Papineni, Roukos, Ward, and
  Zhu}]{acl-papineni:02}
Kishore Papineni, Salim Roukos, Todd Ward, and Wei-Jing Zhu. 2002.
\newblock \uppercase{BLEU}: a method for automatic evaluation of machine
  translation.
\newblock In \emph{Proceedings of the Annual Meeting of the Association for
  Computational Linguistics}.

\bibitem[{Raffel et~al.(2017)Raffel, Luong, Liu, Weiss, and
  Eck}]{raffel2017online}
Colin Raffel, Minh{-}Thang Luong, Peter~J. Liu, Ron~J. Weiss, and Douglas Eck.
  2017.
\newblock \href {http://proceedings.mlr.press/v70/raffel17a.html} {Online and
  linear-time attention by enforcing monotonic alignments}.
\newblock In \emph{Proceedings of the 34th International Conference on Machine
  Learning, {ICML} 2017, Sydney, NSW, Australia, 6-11 August 2017}, volume~70
  of \emph{Proceedings of Machine Learning Research}, pages 2837--2846. {PMLR}.

\bibitem[{Ranzato et~al.(2016)Ranzato, Chopra, Auli, and
  Zaremba}]{iclr-ranzato:16}
Marc'Aurelio Ranzato, Sumit Chopra, Michael Auli, and Wojciech Zaremba. 2016.
\newblock \href {http://arxiv.org/abs/1511.06732} {Sequence level training with
  recurrent neural networks}.
\newblock In \emph{4th International Conference on Learning Representations,
  {ICLR} 2016, San Juan, Puerto Rico, May 2-4, 2016, Conference Track
  Proceedings}.

\bibitem[{Rush et~al.(2015)Rush, Chopra, and Weston}]{rush2015neural}
Alexander~M Rush, Sumit Chopra, and Jason Weston. 2015.
\newblock A neural attention model for abstractive sentence summarization.
\newblock \emph{arXiv preprint arXiv:1509.00685}.

\bibitem[{Salakhutdinov(2014)}]{book-goodfellow:16}
Ruslan Salakhutdinov. 2014.
\newblock \href {https://doi.org/10.1145/2623330.2630809} {Deep learning}.
\newblock In \emph{The 20th {ACM} {SIGKDD} International Conference on
  Knowledge Discovery and Data Mining, {KDD} '14, New York, NY, {USA} - August
  24 - 27, 2014}, page 1973. {ACM}.

\bibitem[{Shang et~al.(2015)Shang, Lu, and Li}]{shang2015neural}
Lifeng Shang, Zhengdong Lu, and Hang Li. 2015.
\newblock \href {https://doi.org/10.3115/v1/P15-1152} {Neural responding
  machine for short-text conversation}.
\newblock In \emph{Proceedings of the 53rd Annual Meeting of the Association
  for Computational Linguistics and the 7th International Joint Conference on
  Natural Language Processing (Volume 1: Long Papers)}, pages 1577--1586,
  Beijing, China. Association for Computational Linguistics.

\bibitem[{Sutskever et~al.(2014)Sutskever, Vinyals, and Le}]{nips-sutskever:14}
Ilya Sutskever, Oriol Vinyals, and Quoc~V. Le. 2014.
\newblock \href
  {https://proceedings.neurips.cc/paper/2014/hash/a14ac55a4f27472c5d894ec1c3c743d2-Abstract.html}
  {Sequence to sequence learning with neural networks}.
\newblock In \emph{Advances in Neural Information Processing Systems 27: Annual
  Conference on Neural Information Processing Systems 2014, December 8-13 2014,
  Montreal, Quebec, Canada}, pages 3104--3112.

\bibitem[{Vaswani et~al.(2017)Vaswani, Shazeer, Parmar, Uszkoreit, Jones,
  Gomez, Kaiser, and Polosukhin}]{nips-Vaswani:17}
Ashish Vaswani, Noam Shazeer, Niki Parmar, Jakob Uszkoreit, Llion Jones,
  Aidan~N. Gomez, Lukasz Kaiser, and Illia Polosukhin. 2017.
\newblock \href
  {https://proceedings.neurips.cc/paper/2017/hash/3f5ee243547dee91fbd053c1c4a845aa-Abstract.html}
  {Attention is all you need}.
\newblock In \emph{Advances in Neural Information Processing Systems 30: Annual
  Conference on Neural Information Processing Systems 2017, December 4-9, 2017,
  Long Beach, CA, {USA}}, pages 5998--6008.

\bibitem[{Vinyals et~al.(2015)Vinyals, Toshev, Bengio, and
  Erhan}]{cvpr-vinyals:15}
Oriol Vinyals, Alexander Toshev, Samy Bengio, and Dumitru Erhan. 2015.
\newblock \href {https://doi.org/10.1109/CVPR.2015.7298935} {Show and tell: {A}
  neural image caption generator}.
\newblock In \emph{{IEEE} Conference on Computer Vision and Pattern
  Recognition, {CVPR} 2015, Boston, MA, USA, June 7-12, 2015}, pages
  3156--3164. {IEEE} Computer Society.

\bibitem[{Wiseman and Rush(2016)}]{emnlp-wiseman:16}
Sam Wiseman and Alexander~M. Rush. 2016.
\newblock \href {https://doi.org/10.18653/v1/D16-1137} {Sequence-to-sequence
  learning as beam-search optimization}.
\newblock In \emph{Proceedings of the 2016 Conference on Empirical Methods in
  Natural Language Processing}, pages 1296--1306, Austin, Texas. Association
  for Computational Linguistics.

\bibitem[{Zhang et~al.(2016)Zhang, Gan, and Carin}]{zhang2016generating}
Yizhe Zhang, Zhe Gan, and Lawrence Carin. 2016.
\newblock Generating text via adversarial training.
\newblock In \emph{Proceedings of the workshop on Adversarial Training,
  Advances in Neural Information Processing Systems}, volume~21.

\bibitem[{Zhou et~al.(2017)Zhou, Yang, Wei, and Zhou}]{zhou2017selective}
Qingyu Zhou, Nan Yang, Furu Wei, and Ming Zhou. 2017.
\newblock Selective encoding for abstractive sentence summarization.
\newblock \emph{meeting of the association for computational linguistics},
  1:1095--1104.

\end{thebibliography}
\bibliographystyle{acl_natbib}

\end{document}